\documentclass[conference,10pt,a4paper]{IEEEtran}
\IEEEoverridecommandlockouts
\usepackage{cite}
\usepackage{amsmath,amssymb,amsfonts}
\usepackage{algorithmic}
\usepackage{graphicx}
\usepackage{textcomp}
\usepackage{xcolor}
\def\BibTeX{{\rm B\kern-.05em{\sc i\kern-.025em b}\kern-.08em
T\kern-.1667em\lower.7ex\hbox{E}\kern-.125emX}}
\begin{document}

\title{EYE-DEX: Eye Disease Detection and EXplanation System}

\author{
  \IEEEauthorblockN{%
    Youssef Sabiri\IEEEauthorrefmark{1}, 
    Walid Houmaidi\IEEEauthorrefmark{1}, 
    Amine Abouaomar
  }
  \IEEEauthorblockA{%
    School of Science and Engineering, Al Akhawayn University\\
    Ifrane, Morocco\\
    Email: \{y.sabiri, w.houmaidi, a.abouaomar\}@aui.ma
  }%
  \thanks{* Youssef Sabiri and Walid Houmaidi contributed equally to this work and are co‑first authors.}
}

\maketitle

\begin{abstract}
Retinal disease diagnosis is critical in preventing vision loss and reducing socioeconomic burdens. Globally, over 2.2 billion people are affected by some form of vision impairment, resulting in annual productivity losses estimated at \$411 billion. Traditional manual grading of retinal fundus images by ophthalmologists is time-consuming and subjective. In contrast, deep learning has revolutionized medical diagnostics by automating retinal image analysis and achieving expert-level performance. In this study, we present EYE-DEX, an automated framework for classifying 10 retinal conditions using the large-scale Retinal Disease Dataset comprising 21,577 eye fundus images. We benchmark three pre-trained Convolutional Neural Network (CNN) models—VGG16, VGG19, and ResNet50—with our fine-tuned VGG16 achieving a state-of-the-art global benchmark test accuracy of 92.36\%. To enhance transparency and explainability, we integrate the Gradient-weighted Class Activation Mapping (Grad-CAM) technique to generate visual explanations highlighting disease-specific regions, thereby fostering clinician trust and reliability in AI-assisted diagnostics.
\end{abstract}


\section{Introduction}
Blindness and visual impairment remain pressing global health challenges. In 2020, an estimated 43 million people were blind worldwide, with an additional 295 million suffering from moderate-to-severe vision impairment. Overall, at least 2.2 billion people live with some form of vision impairment~\cite{WHO2019,Burton2021}. Beyond the human toll, vision loss imposes a staggering socioeconomic burden; recent analyses estimate that unaddressed vision impairment results in an annual global productivity loss of approximately \$411 billion~\cite{Marques2021}. These statistics underscore the urgent need for improved strategies in screening and managing retinal diseases such as diabetic retinopathy (DR), glaucoma, and age-related macular degeneration (AMD).

Traditional diagnosis of retinal diseases relies on manual interpretation of color eye fundus photographs by expert ophthalmologists—a process that is both time-consuming and costly. Studies have reported that a single image can take up to 69$\pm$24 seconds to grade manually~\cite{Yang2022PerformanceOT}, and the inherent subjectivity in human assessment often leads to inter-observer variability. Such limitations hinder large-scale screening efforts, especially in regions with a shortage of specialists.

The advent of artificial intelligence (AI) has introduced transformative opportunities in medical diagnostics. Deep learning, particularly CNNs, has been successfully applied to automate the analysis of retinal fundus images, achieving diagnostic performance that rivals that of clinical experts~\cite{Gargeya2017}. Automated systems can process thousands of images in a fraction of the time required by manual methods, significantly enhancing screening efficiency and potentially reducing the rate of undiagnosed conditions.

Despite these advancements, the clinical adoption of AI is challenged by concerns regarding transparency and trust. Many clinicians remain skeptical of AI systems because their decision-making processes are often opaque, rendering the models as “black boxes” with limited interpretability~\cite{Hanif2021}. This distrust poses a significant barrier to integrating AI into routine diagnostic workflows.

Explainable AI (XAI) techniques, such as Grad-CAM, address these concerns by providing visual explanations of model predictions. Grad-CAM generates heatmaps that highlight the regions within a retinal fundus image that contribute most significantly to the final diagnosis~\cite{Selvaraju2017}. These visualizations help clinicians verify that the AI system focuses on medically relevant features, thereby increasing transparency and fostering trust in automated diagnostic tools.

In this paper, we propose EYE-DEX, a deep learning framework for classifying retinal diseases using eye fundus images and incorporating explainability via Grad-CAM. Our contributions include:
\begin{itemize}
    \item Developing a CNN-based classification framework capable of distinguishing 10 retinal conditions.
    \item Benchmarking three pretrained models (VGG16, VGG19, ResNet50) on the largest public multi retinal disease dataset to date.
    \item Establishing a new state-of-the-art accuracy of 92.36\% on this benchmark dataset.
    \item Integrating Grad-CAM to generate interpretable heatmaps that validate the model’s predictions by localizing disease-specific features.

\end{itemize}
These contributions demonstrate the potential of our approach to deliver rapid, reliable, and transparent diagnostic assistance in ophthalmology.

\section{Literature Review}
\subsection{Role of Medical Imaging in Ophthalmic Diagnosis}
Medical imaging is a cornerstone in modern disease diagnosis, especially in ophthalmology, where high-resolution retinal fundus photography enables detailed visualization of retinal structures and pathology \cite{Bernardes2011DigitalOF}. Eye fundus images capture crucial disease markers (e.g., microaneurysms, optic disc cupping, drusen) that are essential for the early detection of conditions such as DR, glaucoma, and AMD~\cite{Abrmoff2010RetinalIA}. These imaging techniques have significantly improved diagnostic accuracy and facilitated timely clinical interventions, thereby reducing the risk of vision loss.

\subsection{Challenges of Manual Eye Disease Classification}
Despite advances in imaging technology, manual interpretation of retinal fundus photographs remains laborious and subject to variability. Diagnosis relies on expert graders, whose assessments are time-consuming and prone to inter-observer differences~\cite{Pandey2025RealisticFP}. The manual process is not only resource-intensive but also limits screening throughput, particularly in regions with a shortage of ophthalmologists. As a result, there is a growing need for automated systems that can replicate or even exceed human performance in analyzing retinal images.

\subsection{Public Datasets for Retinal Disease Detection}
Early efforts in automated retinal disease detection utilized datasets such as Messidor (with approximately 1,200 images)~\cite{Decencire2014FEEDBACKOA} and the Kaggle EyePACS dataset (over 88,000 images for DR grading)~\cite{Gulshan2016DevelopmentAV}. More recent datasets, like ODIR, have expanded the scope to include multiple diseases, though they often encompass limited class diversity and sample sizes~\cite{HerreraChavez2024MultilabelIC}. In contrast, the Retinal Disease Dataset comprises 5,335 original eye fundus images, annotated into ten distinct classes along with augmented images~\cite{dataset}, thus providing a comprehensive resource that surpasses previous efforts in both scale and diversity.

\subsection{Classical and Deep Learning Methods}
Before the advent of deep learning, classical methods based on hand-crafted features and lesion-specific detectors were employed for retinal fundus analysis~\cite{Alyoubi2020DiabeticRD}. Techniques such as edge detection and template matching formed the basis of early DR screening algorithms. However, these approaches were limited by their dependency on expert-designed features and often lacked generalizability. In recent years, CNNs have become the state-of-the-art for retinal fundus image classification. CNN-based models learn hierarchical representations directly from the pixel data, achieving performance that rivals expert clinicians~\cite{Radha2023UnfoldedDK}. Hybrid methods that combine CNNs with attention mechanisms and transformer architectures have further pushed the envelope, offering improved localization of disease-specific features.

\subsection{Accuracy Trends and Dataset Limitations}
The transition from classical methods to deep learning has led to remarkable accuracy improvements, with state-of-the-art systems reporting AUC values between 0.93 and 1.00 for DR and related conditions~\cite{Gulshan2016DevelopmentAV}. Despite these advances, many models have been evaluated on internal datasets, which may overestimate real-world performance. Common challenges include dataset bias due to non-representative sampling, variability in image acquisition, and severe class imbalance. Such limitations underscore the importance of external validation and robust model design to ensure generalizability across diverse patient populations.

\subsection{Clinical Relevance and Impact}
Automated retinal fundus image analysis holds significant promise in reducing the burden on clinicians by rapidly screening large populations and flagging potential cases for further review. AI-driven systems can provide consistent and objective assessments, thereby facilitating early diagnosis and treatment, which are critical in preventing irreversible vision loss. For instance, the FDA-cleared IDx-DR system demonstrated that deep learning could reliably detect DR, setting a precedent for automated screening tools~\cite{Abrmoff2010RetinalIA}. The integration of such systems in telemedicine and remote monitoring further broadens access to quality eye care, particularly in underserved regions.

\subsection{Retinal Disease Dataset: Scale and Novelty}
The Retinal Disease Dataset represents a significant advance in the field by providing one of the largest collections of eye fundus images for retinal disease classification. With 5,335 original images spanning ten classes (including DR, glaucoma, macular scar, optic disc edema, central serous chorioretinopathy, retinal detachment, retinitis pigmentosa, myopia, and healthy)~\cite{dataset}, this dataset offers unprecedented scale and diversity compared to earlier resources. Its comprehensive annotation and public availability make it an invaluable benchmark for future research. By enabling rigorous evaluation of state-of-the-art models, this dataset lays the groundwork for the development of clinically deployable automated diagnostic systems.

\section{Methods}
\subsection{Proposed System Architecture}
\begin{figure*}[!t]
\centering

\includegraphics[width=0.7\textwidth]{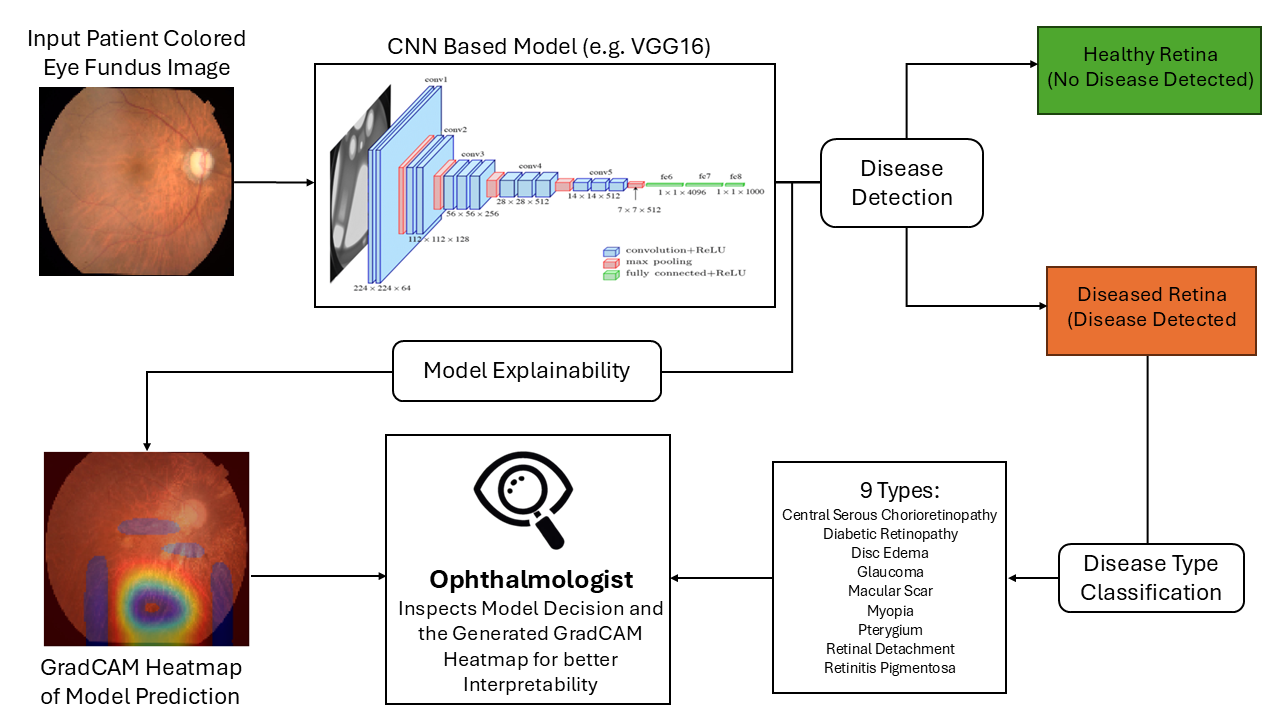}
\caption{Proposed System Architecture of EYE-DEX}
\label{fig:sys_archi}
\end{figure*}

\begin{figure*}[!t]
\centering
\includegraphics[width=0.6\textwidth]{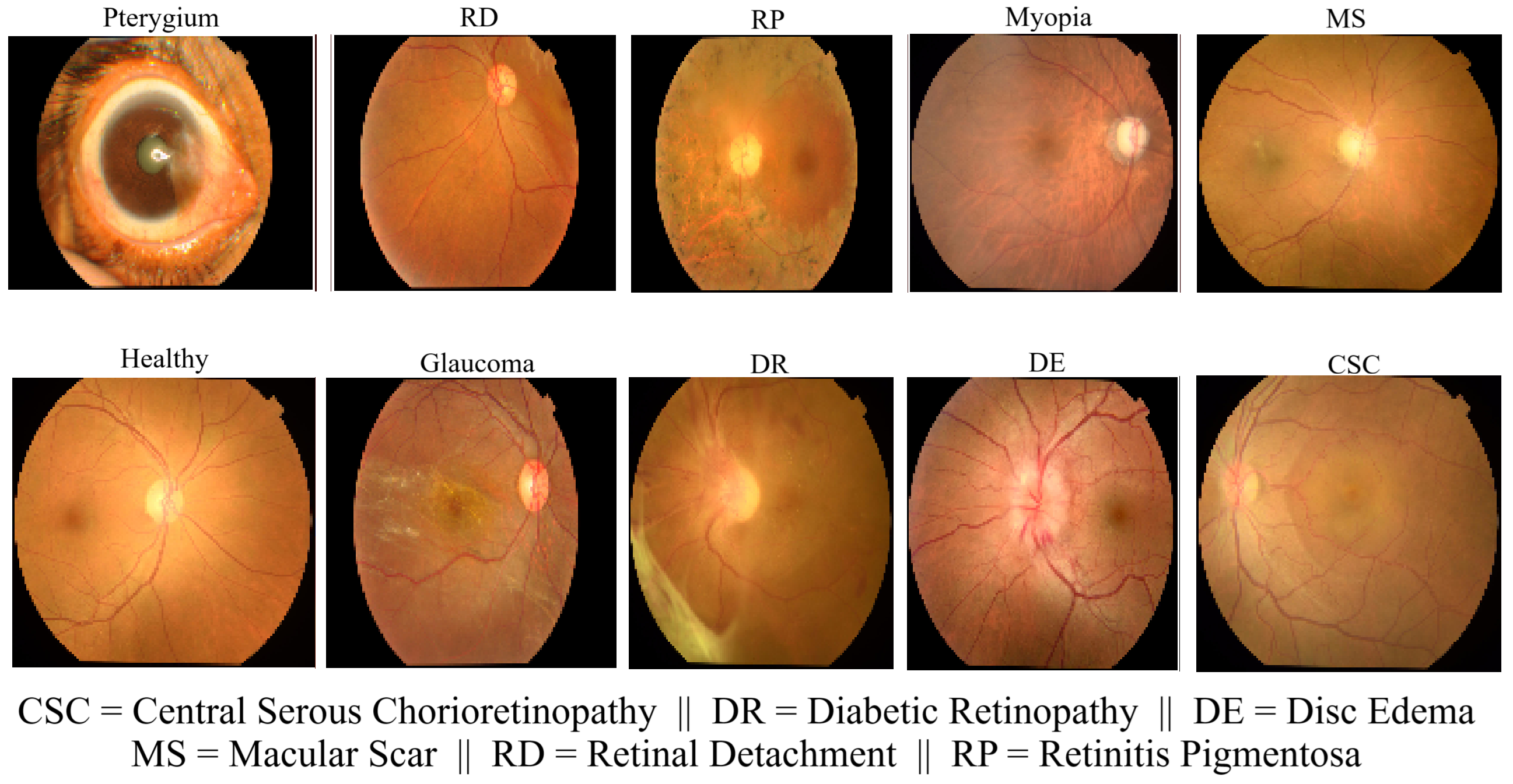}
\caption{Sample Images from the Retinal Disease Dataset, Showcasing the 10 Classes of Retinal Diseases}
\label{fig:dataset_samples}
\end{figure*}
The proposed system architecture of EYE-DEX, illustrated in Figure~\ref{fig:sys_archi}, begins with patient retinal fundus image acquisition, followed by image preprocessing steps (such as resizing and normalization) to ensure consistency across input data. This standardized input is then fed into a CNN, which acts as the primary diagnostic classifier. In the event the CNN detects an anomaly, a subsequent classification stage pinpoints the specific retinal disease among nine possible categories (e.g., DR, glaucoma, macular scar, etc.). Finally, an explainability module employing Grad-CAM is applied to the predicted outcome, generating heatmaps to visually highlight the most influential areas in the retinal image. By integrating these localized explanations, ophthalmologists can verify alignment of the model's focus with clinically relevant features, thus, increasing reliability and transparency in automated detection and classification of retinal diseases.

\subsection{Dataset Description}
The Retinal Disease Dataset \cite{dataset}, employed in this study, represents the largest publicly available dataset for disease classification based on eye fundus images. This dataset comprises 5,335 original retinal fundus images, along with 16,242 augmented images generated through various image transformation techniques. These images capture a range of eye conditions—including Retinitis Pigmentosa, Retinal Detachment, Pterygium, Myopia, Macular Scar, Glaucoma, Disc Edema, Diabetic Retinopathy, and Central Serous Chorioretinopathy—as well as healthy eyes, as shown in Figure \ref{fig:dataset_samples}. The data was collected from multiple eye hospitals and annotated by domain experts, ensuring a high degree of diagnostic accuracy.

\subsection{Data Preprocessing}
Each eye fundus image, originally captured at high resolution, was resized to 224$\times$224 pixels to ensure compatibility with standard deep learning architectures. This resizing not only standardizes the input dimensions for model training but also reduces computational load without sacrificing critical diagnostic information. Additionally, normalization of pixel values was performed, scaling the intensities to the [0, 1] range, which aids in accelerating model convergence and improving overall performance.

\subsection{Dataset Bias}
While the Retinal Disease Dataset comprises a total of 5,335 original images (supplemented by 16,242 augmented images), the distribution among the 10 classes is significantly uneven. Table~\ref{tab:class_imbalance} below shows the original image counts per class.

\begin{table}[h]
\centering
\setlength{\tabcolsep}{6pt} 
\caption{Class Imbalance in the Retinal Disease Dataset}
\begin{tabular}{lcccccc}
\hline
\textbf{Class} & \textbf{Number of Images} \\
\hline
Healthy & 3700 \\
Retinitis Pigmentosa & 973 \\
Retinal Detachment & 875 \\
Pterygium & 119 \\
Myopia & 2751 \\
Macular Scar & 2381 \\
Glaucoma & 4229 \\
Disc Edema & 889 \\
Diabetic Retinopathy & 4953 \\
Central Serous Chorioretinopathy & 707 \\
\hline
\end{tabular}
\label{tab:class_imbalance}
\end{table}

\subsection{Dataset Splits}
The dataset was divided into three subsets: training (80\%), validation (10\%), and testing (10\%). These splits ensure that the performance metrics reflect the model’s ability to generalize beyond the training data.

\subsection{Data Augmentation and Class Balancing}
To address the dataset bias, enhance the generalization capabilities of the model, and mitigate overfitting, we applied data augmentation techniques to the training set. The augmentation strategies included:
\begin{itemize}
\item \textbf{Shear (30\%)}: Introduced random shearing transformations to simulate slight angular variations.
\item \textbf{Zoom (30\%)}: Applied random zooming to mimic different distances or scales.
\item \textbf{Vertical Flips}: Performed vertical flips to create additional variations in the training data.
    
\end{itemize}
These techniques effectively increased the diversity of the training data, thereby helping the model become more robust to variations in real-world imaging conditions. To further mitigate the severe class imbalance, we employed a combination of techniques, including \textbf{class weighting} and \textbf{focal loss}, to give more importance to underrepresented classes during the training process. This approach was preferred over synthetic oversampling methods like SMOTE to avoid the potential for generating noisy or unrealistic data points, which can sometimes degrade model performance on real-world images.

\subsection{Model Selection and Justification}
In our study, we employed several state-of-the-art CNN architectures—VGG16, VGG19, and ResNet50—to robustly classify eye fundus images. Each model was selected for its unique advantages in feature extraction and training stability.
\begin{itemize}
\item \textbf{VGG16:} VGG16 is a well-established CNN that uses small 3$\times$3 filters for effective feature extraction. Its robustness and ease of transfer learning make it an ideal baseline for eye fundus image classification \cite{vgg}.

\item \textbf{VGG19:} VGG19 extends VGG16 with additional layers to capture more complex and subtle image details. Its deeper architecture is beneficial for identifying fine pathological variations in eye fundus images \cite{vgg}.

\item \textbf{ResNet50:} ResNet50 employs residual connections that mitigate the vanishing gradient issue in deep networks. Its robust feature extraction and stable training performance make it well-suited for detailed medical image analysis \cite{resnet}.
    
\end{itemize}

\subsection{Model Fine-Tuning and Architecture}
To adapt the pre-trained CNN architectures to our retinal fundus image classification task, we fine-tuned each model by unfreezing the last 10 layers and appending a custom suite of layers—comprising \texttt{Dropout}, \texttt{BatchNormalization}, \texttt{GlobalAveragePooling2D}, and dense layers with L2 regularization—to enhance feature extraction and mitigate overfitting. We applied these fine-tuning strategies uniformly to all three models to ensure a fair comparison of performance and results.

\subsection{Training Details and Hyperparameters}
\begin{itemize}
\item \textbf{Epochs:} 100 epochs were used with early stopping (patience = 7) to prevent overfitting.
\item \textbf{Learning Rate:} An initial learning rate of 0.0001 was employed, dynamically adjusted via a ReduceLROnPlateau scheduler (halving the rate upon validation loss stagnation, with a minimum of 1e-8).
\item \textbf{Optimizer:} The Adam optimizer was chosen for its efficiency in handling sparse gradients.
\item \textbf{Batch Size:} A batch size of 64 was utilized for model training.
\item \textbf{Loss Function:} Categorical crossentropy was used to train the multi-class classifier.
\item \textbf{Callbacks:} EarlyStopping, ReduceLROnPlateau, and ModelCheckpoint were implemented to monitor validation performance and save the best model.
\item \textbf{Class Imbalance Handling:} To address severe class imbalance, we employed state-of-the-art techniques such as class weighting and focal loss.
\end{itemize}

\subsection{Evaluation Metrics}
To comprehensively assess model performance, we employed several evaluation metrics:
\begin{itemize}
\item \textbf{Accuracy:} Used for overall proportion of correct predictions.
\item \textbf{Precision \& Recall:} Class-specific performance metrics, crucial for handling imbalanced data.
\item \textbf{F1-Score:} Balances precision and recall via their harmonic mean.
\item \textbf{Confusion Matrix:} Visualizes the distribution of correct and incorrect predictions across classes.
\end{itemize}
Collectively, these metrics provide a robust framework for evaluating and comparing the performance of the fine-tuned models, ensuring that both overall performance and class-specific behaviors are thoroughly analyzed.

\section{Results and Discussion}
\subsection{Performance Comparison}
\begin{table}[htbp]
\centering
\setlength{\tabcolsep}{6pt} 
\renewcommand{\arraystretch}{1}
\caption{EYE-DEX: Test Accuracies}
\begin{tabular}{lcccccc}
\hline
\textbf{Model} & \textbf{Test Accuracy (\%)} \\
\hline
VGG16 & \textbf{92.36} \\
ResNet50 & 87.83 \\
VGG19 & 86.89 \\
\hline
\end{tabular}
\label{tab:model_accuracies}
\end{table}

Table~\ref{tab:model_accuracies} clearly compares the accuracies achieved by each model and emphasizes that VGG16 has set a \textbf{new global benchmark} with a test accuracy of \textbf{92.36\%}. It also highlights that this study is the first to evaluate and publish results on this unique, large, multiclass retinal disease image dataset.

\subsection{Results Interpretation}
In our experiments, the fine-tuned VGG16 model achieved the highest accuracy at 92.36\%, establishing a new state-of-the-art for the Retinal Disease Dataset. Its streamlined architecture proved exceptionally well-suited for capturing the subtle pathological features in eye fundus images. In contrast, VGG19’s additional layers did not translate into improved performance, possibly due to a less optimal feature representation for this specific task, leading to an accuracy of 86.89\%. ResNet50, despite its advanced residual connections designed to facilitate training of deeper networks, achieved the second-best test accuracy of 87.83\%, suggesting that its architectural complexity may not be the optimal approach for extracting the distinctive characteristics of these particular retinal images. Overall, these findings indicate that a well-tuned VGG16 provides the most effective balance between model simplicity and feature extraction for this classification task.

\subsection{Confusion Matrix and Classification Report of VGG16 Model}
\begin{figure}[h]
\centering
\includegraphics[width=0.95\columnwidth]{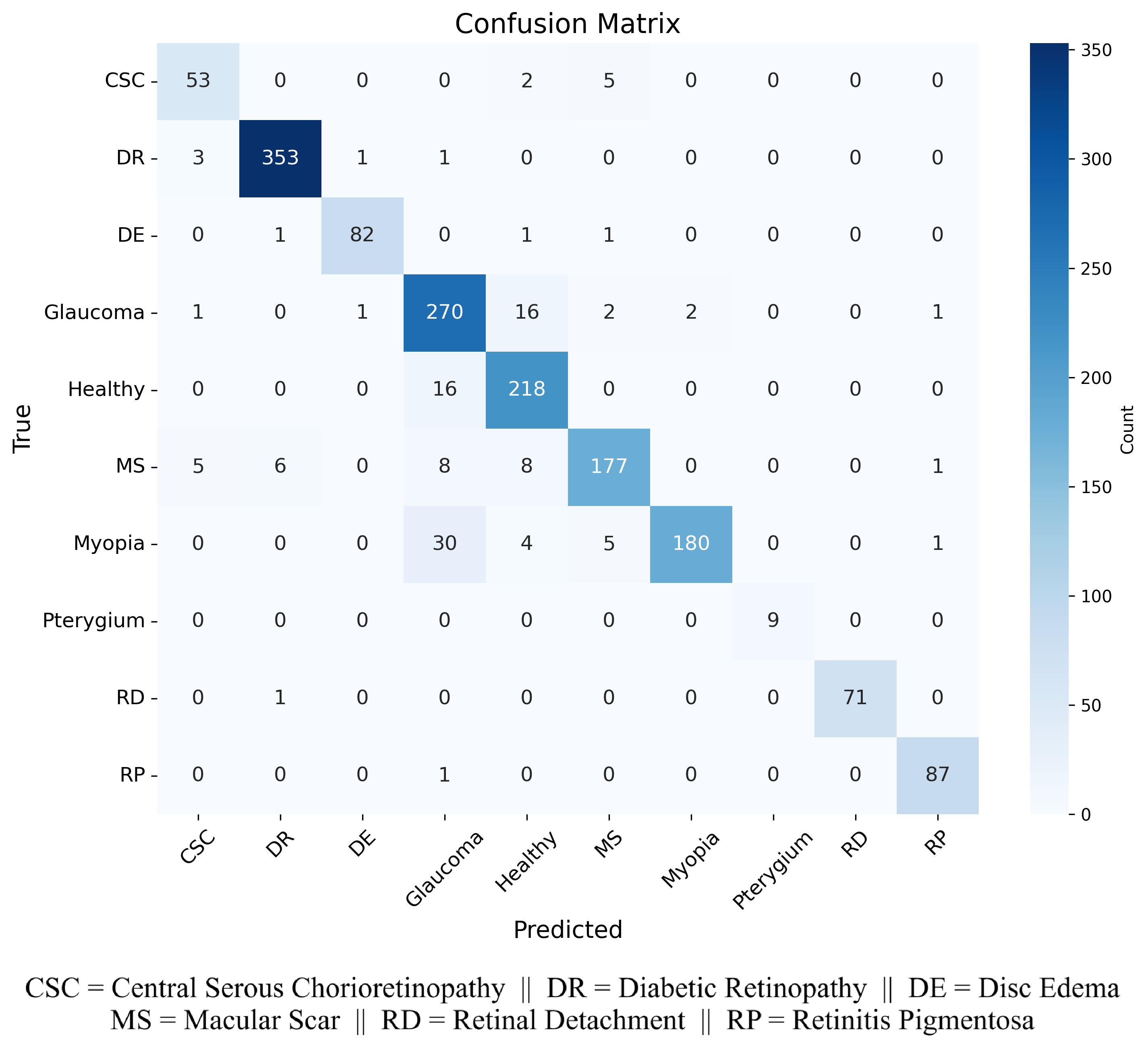}
\caption{Confusion Matrix of VGG16 Model}
\label{fig:confusion_matrix}
\end{figure}

\begin{table}
\centering
\renewcommand{\arraystretch}{1}
\setlength{\tabcolsep}{4pt}
\caption{Detailed Classification Report of VGG16 Model}
\begin{tabular}{lcccc}
\hline
\textbf{Class} & \textbf{Precision} & \textbf{Recall} & \textbf{F1-Score} & \textbf{Support} \\
\hline
Central Serous Chorioretinopathy & 0.85 & 0.88 & 0.87 & 60 \\
Diabetic Retinopathy & 0.98 & 0.99 & 0.98 & 358 \\
Disc Edema & 0.98 & 0.96 & 0.97 & 85 \\
Glaucoma & 0.83 & 0.92 & 0.87 & 293 \\
Healthy & 0.88 & 0.93 & 0.90 & 234 \\
Macular Scar & 0.93 & 0.86 & 0.90 & 205 \\
Myopia & 0.99 & 0.82 & 0.90 & 220 \\
Pterygium & 1.00 & 1.00 & 1.00 & 9 \\
Retinal Detachment & 1.00 & 0.99 & 0.99 & 72 \\
Retinitis Pigmentosa & 0.97 & 0.99 & 0.98 & 88 \\
\hline
\textbf{Accuracy} & & & \textbf{0.92} & 1624 \\
\textbf{Macro Avg} & 0.94 & 0.93 & 0.94 & 1624 \\
\textbf{Weighted Avg} & 0.93 & 0.92 & 0.92 & 1624 \\
\hline
\end{tabular}
\label{tab:classification_report}
\end{table}

Figure~\ref{fig:confusion_matrix} shows the confusion matrix for the best-performing model, while Table~\ref{tab:classification_report} provides a detailed classification breakdown. Overall, the classifier demonstrates strong robustness, achieving a macro-averaged F1-score of 0.94 despite the significant class imbalance in the original dataset. Notably, certain underrepresented classes, such as \emph{Pterygium} (with only 119 original images), achieved perfect scores for both precision and recall. This indicates that the combination of data augmentation, class weighting, and focal loss effectively mitigates bias against rare classes without relying on synthetic data generation. This is a crucial finding, as it suggests our method is more robust to real-world data variations.

Furthermore, classes with higher representation, such as \emph{Diabetic Retinopathy}, maintain exceptionally high metrics (e.g., 0.98 F1-score), confirming that our class balancing approaches did not compromise performance on majority classes. Minor overlaps do occur between visually similar pathologies, like \emph{Myopia} and \emph{Macular Scar}, but the classifier still achieves F1-scores of 0.90 for both. Taken together, these results underscore the efficacy of the adopted strategies for addressing dataset bias, ensuring that both commonly occurring and rare retinal diseases are accurately and consistently identified.

\section{Explainable AI for Retinal Disease Classification}
\begin{figure*}[!t]
\centering

\includegraphics[width=0.7\textwidth]{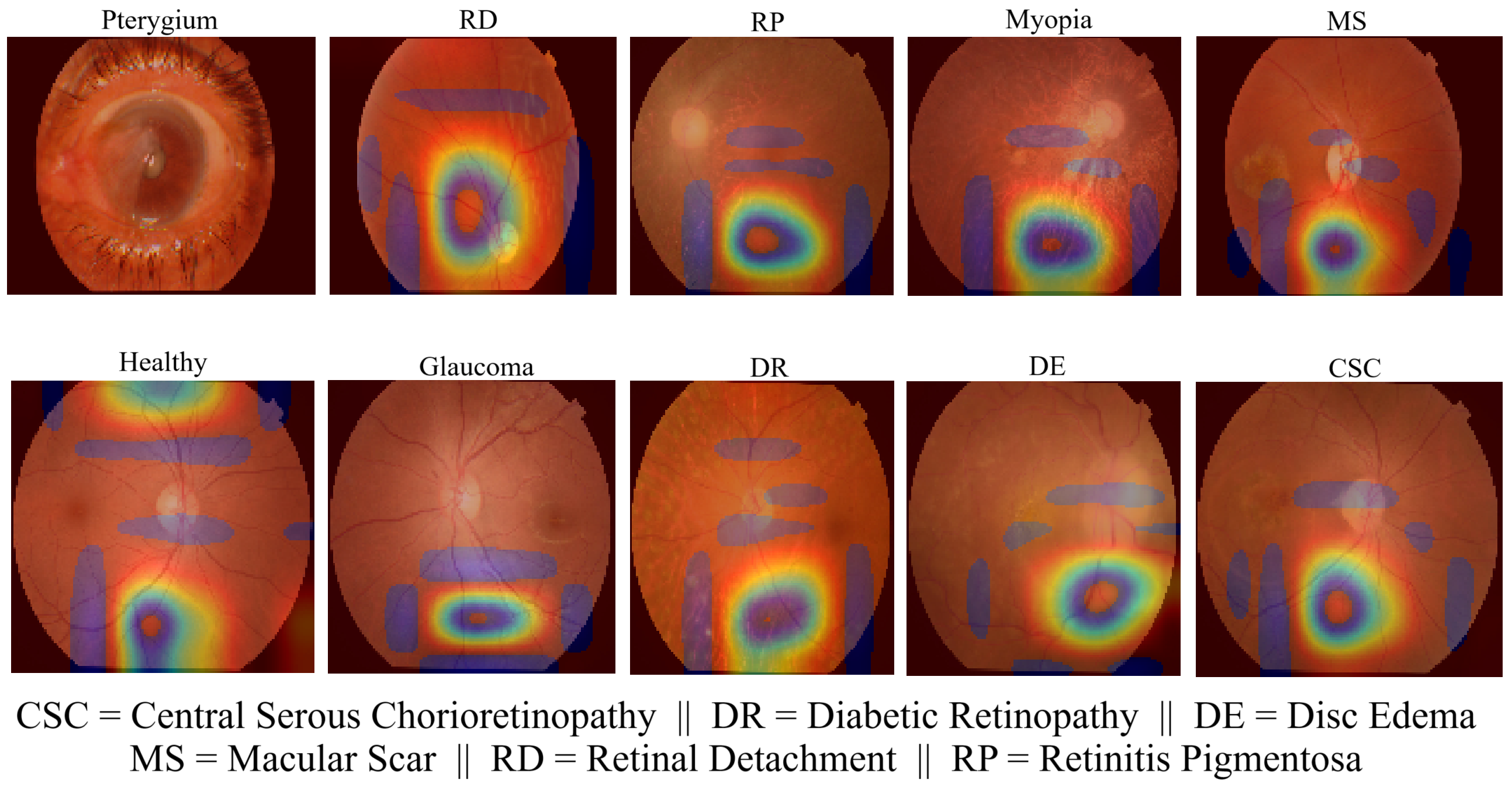}
\caption{Grad-CAM Visualizations for Retinal Disease Classification}
\label{fig:gradcam}
\end{figure*}
XAI plays a crucial role in the EYE-DEX framework by validating our retinal fundus image classification and offering visual insights into the decision-making process of deep learning models. In EYE-DEX, Grad-CAM is employed to generate heatmaps that highlight the specific regions of the eye fundus images which contribute most significantly to the model’s predictions. This approach not only enhances model interpretability but also builds trust among clinicians by confirming that the network focuses on medically relevant features. As illustrated in Figure \ref{fig:gradcam}, the Grad-CAM heatmaps clearly delineate areas of the retina associated with various pathologies, thereby supporting both diagnostic insights and potential clinical validation of the automated classification system.

\section{Conclusion and Future Work}
In summary, this study presents EYE-DEX, a robust deep learning framework that effectively automates the classification of retinal diseases. By leveraging transfer learning with pre-trained CNN models—namely VGG16, VGG19, and ResNet50—the proposed approach not only achieves a state-of-the-art test accuracy of 92.36\% but also demonstrates its capability to discern subtle pathological features within retinal fundus images. The integration of Grad-CAM improves the explainability and interpretability of the model, providing visual insights that build clinician trust by pinpointing disease-specific regions in the images.

The framework also addresses critical challenges such as dataset bias and class imbalance through targeted data augmentation strategies, including focal loss and class weighting. These efforts ensure that both common and rare retinal conditions are accurately identified, paving the way for more reliable and scalable diagnostic tools. Looking ahead, further research may focus on expanding the dataset diversity, incorporating additional imaging modalities, and refining model architectures to further enhance diagnostic performance. Ultimately, this work contributes to the advancement of AI-assisted ophthalmology, offering a promising tool to facilitate early diagnosis and improve patient outcomes while reducing the global socioeconomic burden of vision impairment.

\bibliographystyle{IEEEtran}
\bibliography{references}

\end{document}